\pdfoutput=1

\documentclass[11pt]{article}

\usepackage[final]{acl}

\usepackage{times}
\usepackage{latexsym}
\usepackage{multirow}
\usepackage{graphicx}
\usepackage{booktabs}
\usepackage{enumitem}
\usepackage{algorithm}
\usepackage{algorithmic}
\usepackage{amsmath}
\usepackage{bm}
\usepackage{colortbl}
\usepackage{subcaption}
\usepackage{tabularx}
\usepackage{pifont}

\usepackage[T1]{fontenc}

\usepackage[utf8]{inputenc}

\usepackage{microtype}

\usepackage{inconsolata}

\def\method{\textsc{SemiEvol}}
\title{Semi-supervised Fine-tuning for Large Language Models}

\makeatletter
\renewcommand{\@fnsymbol}[1]{^\dagger}
\makeatother

\author{
Junyu Luo\textsuperscript{\rm $\heartsuit$}, 
Xiao Luo\textsuperscript{\ding{171} \textdagger}, 
Xiusi Chen\textsuperscript{\rm $\diamondsuit$}, 
Zhiping Xiao\textsuperscript{\ding{168} \textdagger},
Wei Ju\textsuperscript{\rm $\heartsuit$}, 
Ming Zhang\textsuperscript{\rm $\heartsuit$ \textdagger} \\
{\textsuperscript{\rm $\heartsuit$} State Key Laboratory for Multimedia Information Processing,} \\
{School of Computer Science, PKU-Anker LLM Lab, Peking University} \\ 
{\textsuperscript{\ding{171}} University of California, Los Angeles}
\quad
{\textsuperscript{\rm $\diamondsuit$} University of Illinois Urbana-Champaign}  
\\
{\textsuperscript{\ding{168}} University of Washington}
\\
{\tt Github Repository: \url{https://github.com/luo-junyu/SemiEvol}.}
}

\def\eg{\emph{e.g}.} 
\def\ie{\emph{i.e}.}

\def\etc{\emph{etc}.}

\definecolor{LightCyan}{rgb}{0.88,1,1}

\newcommand{\paratitle}[1]{\noindent\emph{\textbf{#1}}}

\usepackage{amsmath,amsfonts,bm}
\usepackage{algorithm,amsmath,amssymb,bm,amsthm}
\usepackage{algorithmic}
\usepackage{enumitem}

\newtheorem*{lemma*}{Lemma}

\def\eqref#1{equation~\ref{#1}}

\def\1{\bm{1}}

\DeclareMathAlphabet{\mathsfit}{\encodingdefault}{\sfdefault}{m}{sl}
\SetMathAlphabet{\mathsfit}{bold}{\encodingdefault}{\sfdefault}{bx}{n}

\def\gD{{\mathcal{D}}}

\def\gM{{\mathcal{M}}}
\def\gN{{\mathcal{N}}}

\makeatletter
\def\blfootnote{\xdef\@thefnmark{}\@footnotetext}
\makeatother

\begin{document}
\maketitle

\begin{abstract}
Supervised fine-tuning~(SFT) is crucial in adapting large language models~(LLMs) to a specific domain or task. However, only a limited amount of labeled data is available in practical applications, which poses a severe challenge for SFT in yielding satisfactory results. Therefore, a data-efficient framework that can fully exploit labeled and unlabeled data for LLM fine-tuning is highly anticipated.
Towards this end, we introduce a semi-supervised fine-tuning~(SemiFT) task and a framework named \method{} for LLM alignment from a \textit{propagate-and-select} manner. 
For knowledge propagation, \method{} adopts a bi-level approach, propagating knowledge from labeled data to unlabeled data through both in-weight and in-context methods. 
For knowledge selection, \method{} incorporates a collaborative learning mechanism, selecting higher-quality \textit{pseudo-response} samples. 
We conducted experiments using GPT-4o-mini and Llama-3.1 on seven general or domain-specific datasets, demonstrating significant improvements in model performance on target data. 
Furthermore, we compared \method{} with SFT and self-evolution methods, highlighting its practicality in hybrid data scenarios.
\end{abstract}

\blfootnote{
\textsuperscript{\textdagger} Corresponding authors.
}

\section{Introduction}
Supervised fine-tuning~(SFT) is a crucial method for enhancing large language models'~(LLMs) performance on instructional or domain-specific tasks~\cite{raffel2020exploring,chung2024scaling}, playing a vital role in adapting LLMs for specific scenarios. However, SFT relies on a substantial amount of annotated labeled data, which can be increasingly costly in real-world applications~\cite{honovich2023unnatural,kung2023active}. 
While existing LLMs often employ unsupervised pretraining methods~\cite{devlin2018bert,radford2019language,brown2020language} to improve their capabilities, this approach typically requires vast datasets and substantial computational resources, making it impractical for scenarios with limited accessible samples.
\begin{figure}[t]
    \centering
    \includegraphics[width=0.85\linewidth]{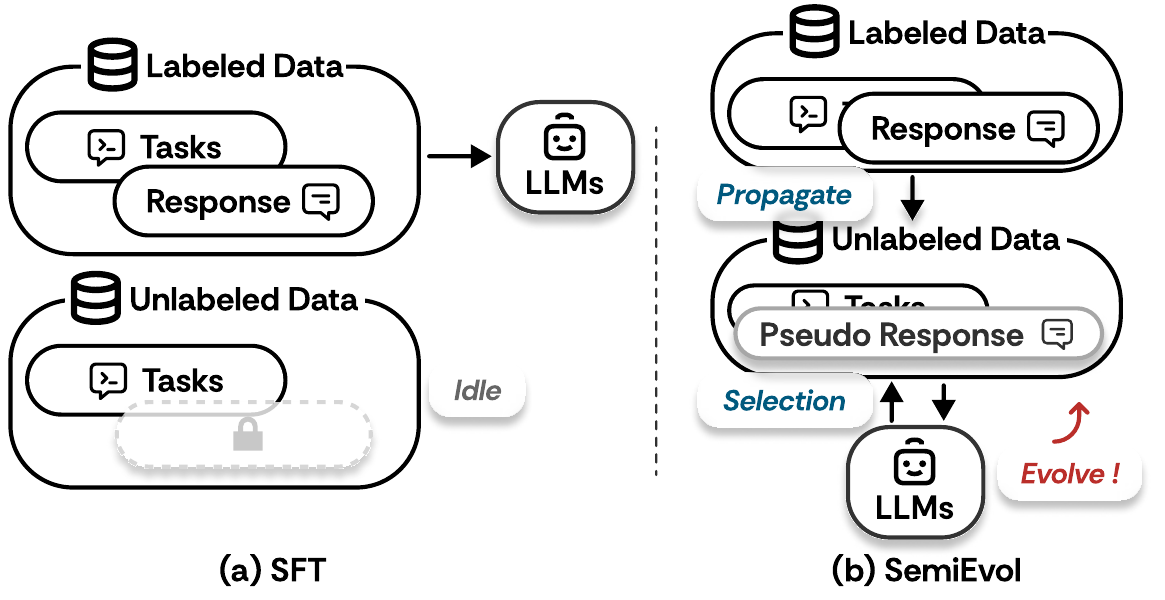}
    \caption{Comparison of \method{} with previous SFT methods. \method{} enables interaction between diverse data types for superior performance evolution.}\label{fig:teaser}
    \vspace{-2mm}
\end{figure}

In practice, however, it often presents a hybrid situation, where a small amount of labeled data coexists with a relatively larger volume of unlabeled data. 
On the one hand, when deploying LLMs to new target tasks, a limited amount of task-specific annotations can be valuable without incurring excessive costs~\cite{perlitz2023active,kung2023active}. 
On the other hand, during the continuous inference process of LLMs, a substantial amount of unlabeled data accumulates~\cite{tao2024survey,honovich2023unnatural,wang2023self}. 
Effectively leveraging the labeled data to enhance model performance on unlabeled data, while simultaneously selecting high-quality unlabeled samples, can improve LLMs' performance in target scenarios, offering substantial practical utility. 
Therefore, we aim to address the following question: 
\begin{quote}
\vspace{-2mm}
\textit{Can LLMs evolve in a real-world scenario of limited labeled data and abundant unlabeled data?}
\vspace{-2mm}
\end{quote}
Designing an evolution framework for hybrid-data scenarios is non-trivial due to the following reasons: First, semi-supervised learning~\cite{kipf2016semi,shi2023rethinking}, which has been widely studied in machine learning, primarily focuses on classification tasks. When considering \textit{generative} tasks, the previous techniques such as pseudo-labeling~\cite{sohn2020fixmatch} and contrastive learning~\cite{he2020momentum}, cannot be directly applied to LLM use cases, like reasoning and planning~\cite{chen2022convfinqa,mmlu}. Second, previous SFT and unsupervised pretraining methods typically deal with a single type of data~(either labeled or unlabeled)~\cite{zhang2023instruction}. Under hybrid-data circumstances, effectively maximizing their combined potential for model improvement becomes challenging.

In this work, we introduce \method{} for improving LLM reasoning in hybrid-data scenarios, as illustrated in Figure~\ref{fig:teaser}. \method{} employs a bi-level strategy for knowledge \textit{propagation-and-selection}. For knowledge propagation, \method{} enhances LLMs' inference performance using labeled data through both in-weight and in-context scopes. During in-weight propagation, \method{} uses labeled data to adapt the model. During in-context propagation, \method{} employs k-nearest neighbor retrieval in latent space to assist prediction.
Moreover, \method{} introduces a bi-level approach for data selection and generating \textit{pseudo-responses}. First, it introduces a collaborative learning framework, utilizing multiple LLMs with different configurations for inference and self-justification of responses, yielding more accurate predictions. Second, \method{} adaptively selects unlabeled data by confidence based on response entropy.
By mining on unlabeled data leveraging labeled data, we obtain high-quality \textit{pseudo-responses}. Using these \textit{pseudo-response} data, the model enhances its performance on target tasks. We conducted tests on seven general or domain-specific datasets~(\eg, MMLU, MMLU-Pro and ConvFinQA), covering tasks such as question-answering, reasoning, and numerical computation. We compared \method{} with popular methods like retrieval augmented generation, self-evolution and SFT, demonstrating \method{}'s consistent effectiveness across various scenarios.

We summarize the contributions as follows:
\begin{itemize}[leftmargin=*]
\vspace{-2mm}
\item To the best of our knowledge, we are the first to study a practical problem of \textbf{semi-supervised fine-tuning~(SemiFT)}, aiming to adapt LLMs into different domains data-efficiently.  
\vspace{-2mm} 
\item We introduce \method{}, a unified framework for knowledge \textit{propagation-and-selection} that effectively combines labeled and unlabeled data for model evolution.
\vspace{-2mm}
\item We demonstrate the consistent effectiveness of \method{} across seven widely used general or domain-specific generative tasks in comparison to extensive baseline models.
\vspace{-2mm}
\end{itemize}


\section{Challenges for Real-world LLM Fine-tuning}

\begin{figure*}[t]
    \centering
    \includegraphics[width=\textwidth]{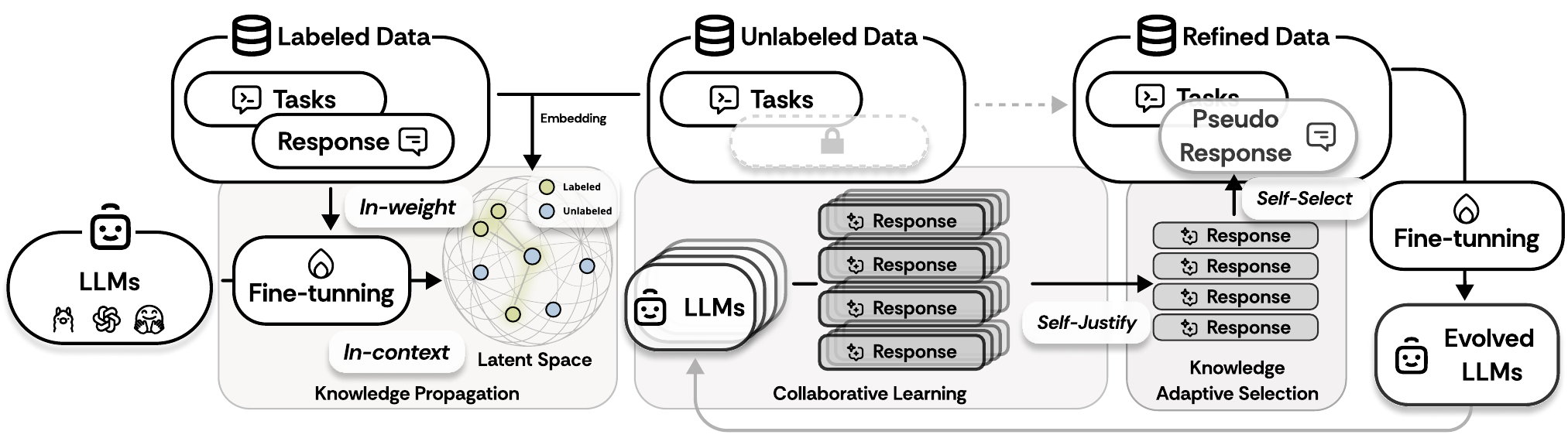}
    \vspace{-4mm}
    \caption{Overview of \method{}. It maximizes the utility of labeled data through a bi-level knowledge \textit{propagation-and-selection} framework, while leveraging collaborative learning among multiple LLMs to exploit unlabeled data, thereby unleashing the full data potential. }
    \label{fig:framework}
    \vspace{-4mm}
\end{figure*}

\subsection{Supervised Fine-tuning}

Supervised fine-tuning~(SFT) aims to adapt Large Language Models~(LLMs) to domain-specific scenarios. Given an LLM $\gM$ and a dataset $\gD_{\text{labeled}}=\{T_i, Y_i\}^N_{i=1}$, where $T_i$ represents the input task or context and $Y_i$ denotes the corresponding expected response. The model minimizes the loss function for each token of the anticipated output during the fine-tuning process $FT$.

\textbf{Challenge: Annotation Cost.} Despite the effectiveness of supervised fine-tuning, it would require expensive labeling costs to access abundant labeled data. An economic solution is to utilize easily accessible unlabeled data without feedback as a supplement for fine-tuning.  

\subsection{Background and Problem Definition: Semi-supervised Fine-tuning~(SemiFT)}

In real-world scenarios, it is more common to have access to both a small amount of labeled data $\gD_{\text{labeled}}$ and a larger volume of unlabeled data $\gD_{\text{unlabeled}}=\{T_i\}^M_{i=1}$. Labeled data offers higher confidence, while unlabeled data represents a broader sample distribution. In this paper, we propose \method{} approach, which primarily focuses on how to leverage both types of data $\gD_{\text{semi}} = \gD_{\text{labeled}}~\cup~\gD_{\text{unlabeled}}$ to optimize the LLM $\gM$. Our \method{} not only improves model performance but also offers greater practical value.

\textbf{Challenge: Generative Task.} In fact, developing a semi-supervised fine-tuning framework is highly challenging. Tradition semi-supervised approaches usually focus on classification problems solved by pseudo-labeling while our problem is a generative task, which requires us to generate expected responses instead. 

\section{Methodology}

\subsection{Overview}

In this paper, we develop \method{} to integrate labeled and unlabeled data for improving LLM performance in reasoning. 
The core idea of \method{} is to leverage labeled data through a bi-level  \textit{propagation-and-select} process.
As illustrated in Figure~\ref{fig:framework}, \method{} is featured by three key components:
\textit{(1) Knowledge Propagation}: 
We utilize labeled data to enhance model $\gM$'s performance on unlabeled data. This process focuses on two aspects, \ie, model weights and context. The propagation process involves model adaptation using labeled data and providing the most relevant references from the latent space to assist model inference.
\textit{(2) Collaborative Learning}: We employ multiple LLMs with different configurations as mutual teachers to infer unlabeled data. We pay particular attention to inconsistent responses, using the models to self-justify these discrepancies.
\textit{(3) Knowledge Self-selection}: We design the adaptive selection for unlabeled data and pseudo-responses. Using labeled data as a guide, we identify the most valuable unlabeled data for learning. By optimizing LLMs on these selected data samples, the model achieves superior evolution performance.

In summary, \method{} addresses the prevalent real-world scenario where both labeled and unlabeled data coexist.
By leveraging the labeled data and the capabilities of LLMs themselves, we perform knowledge propagation, mining, and selection on unlabeled data. 
This strategy improves model performance in the target scenarios.

\subsection{Knowledge Propagation}

Labeled data contain expected target responses, while unlabeled data represents a broader task distribution. To leverage this, we aim to propagate knowledge from labeled to unlabeled data, enabling the model to effectively utilize and learn from unlabeled instances. We design a bi-level knowledge propagation framework that operates simultaneously on two fronts: \textit{in-weight} and \textit{in-context}. 

\textit{For in-weight propagation}, we initially warm up the base model $\gM_{base}$ on labeled data $\gD_{\text{labeled}}$ to enhance its predictive capabilities for the target task. Specifically, we fine-tune the model, leveraging task data and target responses to obtain a preliminary adapted model~($\gM_{warm}$). This process is formulated as:
\begin{equation}\label{eq:warm-up}
\gM_{warm} ~=~ FT\left(\gM_{base}, \gD_{\text{labeled}} \right)\,,
\end{equation}
where $FT$ is the fine-tuning process.

\textit{For in-context propagation}, we first embed labeled dataset into latent space using an embedding function $\epsilon(\cdot)$:
\begin{equation}
E_{\text{labeled}} = \left\{ \epsilon \left(t_i\right) ~\lvert~ \left(t_i, y_i\right) \in \gD_{\text{labeled}} \right\} \,.
\end{equation}
During inference on unlabeled data, for each task $t_j \in \gD_{\text{unlabeled}}$, we retrieve the $k$ nearest labeled instances in the embedding space:
\begin{equation}
\gN\left( t_j \right) = {NN}\left( E_{\text{labeled}}, \epsilon\left(t_j\right), k \right)\,,
\end{equation}
where $k$ is set to $3$, $NN$ is the nearest neighbors search. We use $\gN\left( t_j \right)$ as context to improve the inference on the unlabeled data.

In summary, labeled data facilitates knowledge propagation to unlabeled data through both in-weight and in-context manners. In practice, we first adapt the model to obtain the warm-up LLM $\gM_{warm}$, then utilize labeled data as context to enhance inference on unlabeled instances.

\subsection{Collaborative Learning}

To further exploit unlabeled data, we designed a collaborative learning framework tailored for LLMs. This framework utilizes the inherent capabilities of LLMs for \textit{self-justify} to obtain high-confidence \textit{pseudo-responses} from unlabeled data. 
Some concurrent works also attempt to use LLMs for similar functionality~\cite{wang2024self}, while their focus differs from ours.

Initially, we employ a set of $n$ LLMs, denoted as $\gM_1, \gM_2, \cdots, \gM_n$ to perform inference on the unlabeled dataset $\gD_{\text{unlabeled}}$, where $n$ is $4$ by default and will be discussed in Section~\ref{sec:sen_anal}. 
Each model is configured with different inference contexts and settings, providing diverse perspectives and yielding more comprehensive results. For each unlabeled sample 
$t_j\in\gD_{\text{unlabeled}}$, we obtain multiple predictions:
\begin{equation}\label{eq:multi-pred}
\left\{ y^m_j\right\} = \left\{\gM_m \left( t_j \right) \right\}^n_{m=1} \,.
\end{equation}

Subsequently, we implement a \textit{self-justification} process using LLMs. This step synthesizes the inferences from various models to select and summarize the most accurate response $\hat{y_j}$ :
\begin{equation}\label{eq:self-justify}
\tilde{y_j} = \text{Self-Justify}\left( \left\{ y^m_j \right\}^n_{m=1} \right)\,.
\end{equation}
where the \textit{Self-Justify} operator is implemented via prompting $\gM_{warm}$ by natural language instructions.
In summary, our LLM-specific collaborative learning framework harnesses multiple differently configured LLMs for multi-perspective inference. By utilizing the LLMs' inherent abilities to \textit{self-justify}, we effectively mine unlabeled data, and generate high-confident pseudo-responses.

\subsection{Knowledge Adaptive Selection}

While the \textit{pseudo-responses} $\tilde{y}_j$ generated through the collaborative learning framework enrich the training data, they may still contain noise or low-quality information that could misguide the model's learning. To address this issue, we design an adaptive data selection approach within the \method{} framework. 
Specifically, we measure the confidence of the responses $\tilde{y}_j$ for the unlabeled data selection.

We use the entropy of the LLM's responses to measure the model's confidence in the answers.
Since LLMs generate responses token by token, we calculate the per-token negative log-likelihood, which serves as an approximation of the entropy. 
For each data sample $t_j \in \gD_{\text{unlabeled}}$, the entropy $H\left(\tilde{y}_j\right)$ is computed on pseudo-response $\tilde{y}_j$ after Eq.~\ref{eq:self-justify} as:
\begin{equation} 
H\left(\tilde{y}_j\right) = - \frac{1}{L_j} \sum_{k=1}^{L_j} \log P\left(r_j^k \mid t_j, r_j^{<k}\right)\,, 
\end{equation} 
where $L_j$ is the length of the response $r_j$ generated by $\gM_{warm}$, $r_j^k$ is the $k$-th token in the response, $r_j^{<k}=\left\{ r^1_j, r^2_j, \cdots ,r^k_j \right\}$ are the preceding tokens of $\tilde{y}_j$, and $P\left(r_j^k \mid t_j, r_j^{<k}\right)$ is $\gM_{warm}$'s predicted probability of token $r_j^k$ at position $k$.

For the unlabeled data, we compute the entropy $H\left(\tilde{y}_j\right)$ for each pseudo-response $\tilde{y}_j$ corresponding to task $t_j \in \gD_{\text{unlabeled}}$. We then use the $\theta$ percentile of the entropy values from the labeled data to establish a dynamic threshold $\tau$:
\begin{equation} 
\tau = \text{Percentile}_{\theta}\left( \left\{ H\left(\tilde{y}_j\right) \right\}_{j=1}^{M} \right)\,, 
\end{equation} 
where $M$ is the amount of unlabeled samples, and $\theta$ is default to $50\%$ and will be investigated in Section~\ref{sec:sen_anal}.

Using this dynamic threshold, we select confident samples from the unlabeled data. In formula, 
\begin{equation} \label{eq:data-selected}
\gD_{\text{selected}} = \left\{ \left(t_j, \tilde{y}_j\right) ~|~ H\left(\tilde{y}_j\right) \leq \tau \right\}\,. 
\end{equation}
We filter the pseudo-responses obtained previously, resulting in the refined dataset $\gD_{\text{selected}}$.

Finally, we combine the selected pseudo-labeled data with the original labeled data to fine-tune the base model, which can enhance its performance and adaptability on the target task:
\begin{equation} 
\gM_{\text{evol}} = FT\left( \gM_{\text{base}}, \gD_{\text{selected}}~\cup~\gD_{\text{labeled}} \right)\,, 
\end{equation}
where $\gM_{\text{base}}$ is the pre-trained LLM, and $FT$ denotes the fine-tuning process.

By leveraging both high-quality pseudo-labeled data and original labeled data, we enhance the model's performance and adaptability on the target task while reducing the influence of noisy or erroneous information.

\subsection{Summary}

\method{} enhances the performance and adaptability of LLMs in target tasks through a two-stage knowledge mining process, combining labeled and unlabeled data for model evolution. Firstly, we leverage a small amount of labeled data to enhance knowledge propagation across unlabeled data. Secondly, we employ knowledge mining and adaptive selection. This strategy effectively integrates both labeled and unlabeled data, culminating in the evolved model $\gM_{\text{evol}}$. 

\begin{table*}[t]
\centering
\renewcommand{\arraystretch}{0.9}
\setlength{\tabcolsep}{4pt}
\resizebox{.9\linewidth}{!}{
\begin{tabular}{lcccccccc}
\toprule
\textbf{Model and Strategy} & \textbf{MMLU} & \textbf{MMLU Pro} & \textbf{ARC} & \textbf{FPB} & \textbf{USMLE} & \textbf{PubMedQA} &  \textbf{ConvFinQA} \\
\midrule
GPT-4o-mini & \multirow{2}{*}{77.4}  & \multirow{2}{*}{57.8} & \multirow{2}{*}{91.5} & \multirow{2}{*}{93.4} & \multirow{2}{*}{73.8} & \multirow{2}{*}{77.5} & \multirow{2}{*}{63.9} \\
\textit{Vanilla} & & & & & & & \\
\addlinespace[0.5mm]
GPT-4o-mini & \multirow{2}{*}{77.8} & \multirow{2}{*}{58.8} & \multirow{2}{*}{90.3} & \multirow{2}{*}{98.0} & \multirow{2}{*}{75.0} & \multirow{2}{*}{77.5} & \multirow{2}{*}{88.8} \\
\textit{SFT} & & & & & & & \\
\addlinespace[0.5mm]
\rowcolor{gray!10} GPT-4o-mini & & & & & & & \\
\rowcolor{gray!10} \textbf{\method{}} & \multirow{-2}{*}{\textbf{79.9}} & \multirow{-2}{*}{\textbf{60.8}} & \multirow{-2}{*}{\textbf{92.7}} & \multirow{-2}{*}{\textbf{98.9}} & \multirow{-2}{*}{\textbf{77.2}} & \multirow{-2}{*}{\textbf{79.5}} & \multirow{-2}{*}{\textbf{89.2}} \\
\addlinespace[0.5mm]
\midrule
Llama3.1-8B & \multirow{2}{*}{66.4} & \multirow{2}{*}{47.1} & \multirow{2}{*}{81.1} & \multirow{2}{*}{81.7} & \multirow{2}{*}{70.2} & \multirow{2}{*}{73.5} & \multirow{2}{*}{51.1} \\
\textit{Vanilla} & & & & & & & \\
\addlinespace[0.5mm]
Llama3.1-8B & \multirow{2}{*}{67.9} & \multirow{2}{*}{49.8} & \multirow{2}{*}{81.8} & \multirow{2}{*}{96.2} & \multirow{2}{*}{70.8} & \multirow{2}{*}{75.0} & \multirow{2}{*}{81.3} \\
\textit{SFT} & & & & & & & \\
\addlinespace[0.5mm]
AdaptLLM & -- & -- & -- & 49.7 & 31.5 & 27.6 & 30.9 \\
\addlinespace[0.5mm]
InstructPT & -- & -- & -- & 76.1 & 47.4 & 44.5 & 55.2 \\
\addlinespace[0.5mm]
MemoryLLM & 56.4 & 31.8 & 56.3 & 57.7 & 37.8 & 55.5 & 37.2 \\
\addlinespace[0.5mm]
RAG~(BM25) & 66.6 & 37.4 & 80.8 & 83.7 & 69.3 & 69.0 & 63.4 \\
\addlinespace[0.5mm]
RAG~(FAISS) & 66.5 & 38.8 & 81.3 & 82.5 & 69.1 & 71.5 & 64.6 \\
\addlinespace[0.5mm]
Hermes-3& 63.6 & 37.9 & 74.9 & 73.9 & 54.5 & 68.5& 54.9 \\
\addlinespace[0.5mm]
Reflection-Llama& 65.5 & 37.5 & 82.2 & 80.8 & 67.4 & 77.5 & 40.8 \\
\addlinespace[0.5mm]
\rowcolor{gray!10} Llama3.1-8B & & & & & & & \\
\rowcolor{gray!10} \textbf{\method{}} & \multirow{-2}{*}{\textbf{68.8}} & \multirow{-2}{*}{\textbf{50.3}} & \multirow{-2}{*}{\textbf{83.4}} & \multirow{-2}{*}{\textbf{96.2}}  & \multirow{-2}{*}{\textbf{71.6}} & \multirow{-2}{*}{\textbf{76.0}} & \multirow{-2}{*}{\textbf{82.4}} \\
\addlinespace[0.5mm]

\bottomrule
\end{tabular}
}
\caption{\textbf{Performance comparison} across different models on various datasets.}
\label{tab:model_comparison}
\end{table*}

\section{Experiment}

\subsection{Experiment Setup}

\subsubsection{Datasets}

We employed both general-purpose and domain-specific evaluation datasets to provide a comprehensive assessment. These datasets encompass a variety of tasks, including multiple-choice questions, reasoning, numerical computations, \etc. Specifically, our general evaluation datasets include MMLU~\cite{mmlu}, MMLU-Pro~\cite{mmlupro}, and ARC~\cite{clark2018think}, while domain-specific datasets comprise FPB~\cite{malo2014good}, USMLE~\cite{jin2021disease}, PubMedQA~\cite{jin2019pubmedqa,jiang2024tc}, and ConvFinQA~\cite{chen2022convfinqa}, covering various fields such as finance and healthcare. This diverse selection enables a thorough evaluation of the model's performance across different task types and knowledge domains.

\subsubsection{Backbones and Baselines}

\paratitle{Base Models.} To demonstrate the generalization capability of \method{}, we employed a diverse range of leading models, encompassing both commercial and open-source and LLMs, including GPT-4o-mini and Llama-3.1-8B~\cite{dubey2024llama}.

\paratitle{Baselines.} We evaluated our method against baselines from several categories: 
\textit{(1) Vanilla}, which involves testing solely through API calls or using the original model; 
\textit{(2) Supervised Fine-tuning}~(SFT)~\cite{hu2021lora,wei2021finetuned}, which adapts the model to the target task using the labeled data; 
\textit{(3) Self-Evolution Methods}~(SelfEvol), which enhance LLM capabilities using additional unlabeled data. We compare with Reflection-Llama~\cite{li-etal-2024-selective}\footnote{https://huggingface.co/Solshine/reflection-llama-3.1-8B} and Hermes-3~\cite{teknium2024hermes3technicalreport}\footnote{https://huggingface.co/NousResearch/Hermes-3-Llama-3.1-8B}, both of which evolve from the Llama-3.1-8B model; 
\textit{(4) Domain Adaptation Methods}, including AdaptLLM~\cite{cheng2024adapting} and InstructPT~\cite{cheng2024instruction}, utilize domain-specific data~(\eg, finance and medical). We select models adapted to corresponding domains for testing, all with comparable parameter counts of 8B;
\textit{(5) Inference-time enhancement methods}, such as Retrieval Augmented Generation~(RAG)~\cite{lewis2020retrieval}, including BM25~\cite{jones2000probabilistic} and FAISS~\cite{douze2024faiss} algorithms. We also compare with MemoryLLM~\cite{memoryllm}, with the nearest labeled sample as memory; 

This comprehensive comparison allows us to assess the effectiveness of our proposed method across various state-of-the-art approaches in LLM fine-tuning and adaptation.

\subsubsection{Implementation Details}

For the setting of semi-supervised fine-tuning of LLMs, we have $\gD_{\text{labeled}}$, $\gD_{\text{unlabeled}}$ and $\gD_{\text{test}}$. The data proportion in our experiments is $labeled:unlabeled:test = 2:6:2$ and will be further discussed in Section~\ref{sec:data-ratio}.  The answer information for $\gD_{\text{unlabeled}}$ is inaccessible in our setting. We fine-tuned Llama-3.1-8B using Low-Rank Adaptation~(LoRA)~\cite{hu2021lora} and applied fine-tuning with the official API for GPT-4o-mini\footnote{https://platform.openai.com/finetune.}. All fine-tuning processes take $2$ epochs. $n$ is set to $4$ and $\theta$ is set to $50\%$, with further investigation planned in subsequent experiments. Our dataset is publicly available at Hugging Face\footnote{\url{https://huggingface.co/datasets/luojunyu/SemiEvol}}.
 

We evaluated all methods using the test sets $\gD_{\text{test}}$. Model inference followed default settings for each approach. Codes are available in our GitHub repository\footnote{\url{https://github.com/luo-junyu/SemiEvol}}.

\subsection{Main Result}

We present the main results of \method{} in Table~\ref{tab:model_comparison}. We can draw the following insights. 
\textbf{Firstly, the tasks are generally challenging.} Off-the-shelf LLMs perform poorly on these tasks, highlighting the necessity of leveraging scenario data to enhance model performance.
\textbf{Secondly, \method{} consistently improves both commercial and open-source models.} Notably, \method{} is one of the few approaches that demonstrably enhances state-of-the-art commercial models, underscoring its practical value.
\textbf{Thirdly, SFT yield modest improvements,} demonstrating the effectiveness of labeled data. Given the high cost of data labeling, \method{} effectively utilizes unlabeled data to complement this approach.
\textbf{Fourthly, the self-evolution method fails to achieve consistent improvements,} showing limited improvement or even adverse effects on most datasets.
\textbf{Fifthly, adaptive fine-tuning methods can enhance performance only on specific tasks}~(\eg, ConvFinQA). Also, these methods may compromise the model's instruction-following ability, leading to significant performance drops in some tasks~(\eg, USMLE and PubMedQA).
\textbf{Lastly, \method{} consistently outperforms SFT methods}, which demonstrates the effectiveness of incorporating unsupervised data and leveraging labeled data to fully utilize unsupervised data. Even when base models perform poorly~(\eg, MMLU-Pro and ConvFinQA), \method{} can still achieve substantial improvements in model performance.

\subsection{Analysis and Discussions}

\subsubsection{Ablation Study}

To evaluate the effectiveness of different components, we conducted an ablation analysis on \method{}, with results presented in Table~\ref{tab:variant_comparison}. The findings reveal several key insights:
\textbf{(1)} The full model consistently outperforms all other configurations across the three datasets, demonstrating its comprehensive effectiveness.
\textbf{(2)} In terms of knowledge propagation, both In-weight Propagation~(IWP) and In-context Propagation~(ICP) contribute significantly to the transfer of knowledge from labeled to unlabeled data and subsequent model evolution. In-weight Propagation, in particular, shows a more pronounced impact.
\textbf{(3)} Removing Collaborative Learning~(CL) negatively affects model performance. This suggests that Collaborative Learning effectively leverages predictions from multiple LLMs to autonomously identify more accurate answers, thereby enhancing the prediction quality on unlabeled data.
\textbf{(4)} The absence of Adaptive Selection~(AS) also leads to decreased model performance. This indicates that AS successfully selects more confident samples, thus improving the accuracy of unlabeled data and enhancing the model's evolutionary process.

\begin{table}[t]
\centering
\resizebox{.95\columnwidth}{!}{
\begin{tabular}{rccc}
\toprule
\textbf{Variant} & \textbf{MMLU} & \textbf{MMLU-Pro} & \textbf{ARC} \\
\midrule
\rowcolor{gray!10} Llama3.1-8B & & & \\
\rowcolor{gray!10} \quad \textbf{\method{}} & \multirow{-2}{*}{68.8} & \multirow{-2}{*}{50.3} & \multirow{-2}{*}{83.4} \\
\quad \textit{w/o} IWP & 67.8 & 48.8 & 82.4 \\
\quad \textit{w/o} ICP & 68.0 & 49.4 & 83.0 \\
\quad \textit{w/o} CL & 67.0 & 49.2 & 82.4 \\
\quad \textit{w/o} AS & 67.9 & 49.0 & 82.1 \\
\bottomrule
\end{tabular}
}
\caption{\textbf{Ablation study} via performance comparison of different variants on \method{}.}
\label{tab:variant_comparison}
\end{table}

\begin{figure}[t] 
    \centering 
    \begin{subfigure}[b]{0.49\columnwidth} 
        \includegraphics[width=\textwidth]{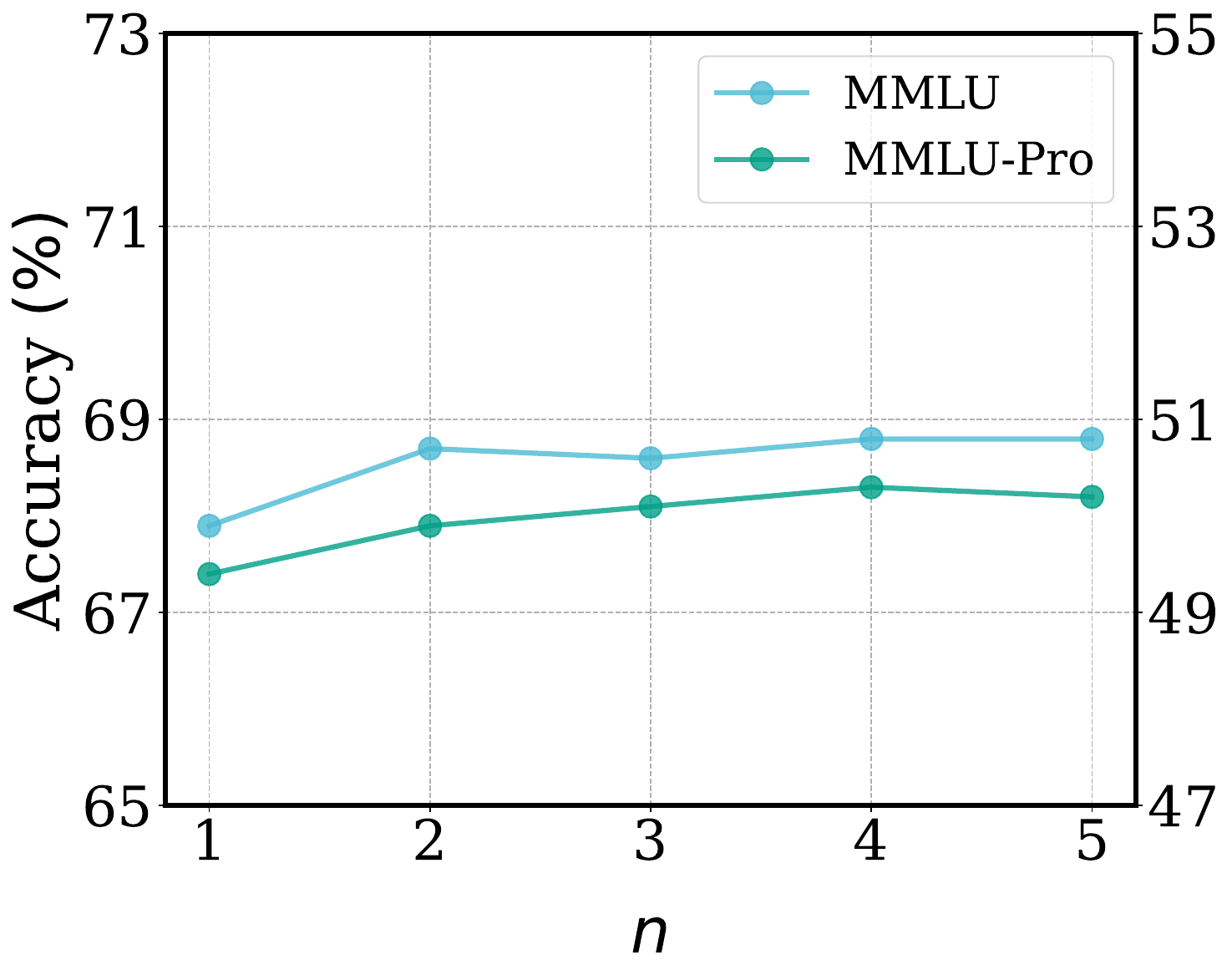}
    \end{subfigure}
    \hfill
    \begin{subfigure}[b]{0.49\columnwidth}
        \includegraphics[width=\textwidth]{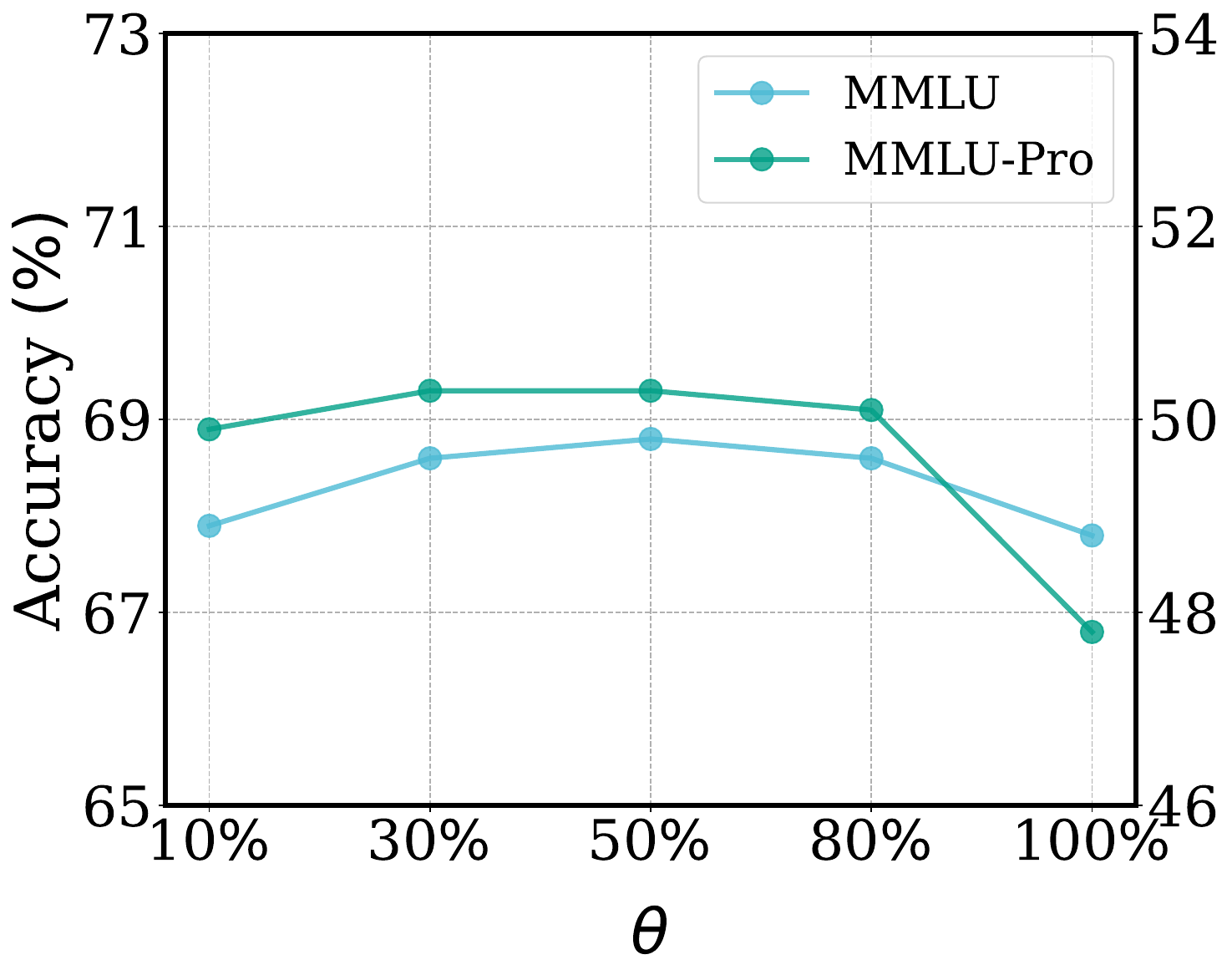} 
    \end{subfigure}
\caption{\textbf{Sensitivity analysis} of \method{}'s performance under different $n$ and $\theta$ on variant datasets.}
\label{fig:sensitive} 
\end{figure}

\begin{figure}[t] 
    \centering 
    \begin{subfigure}[b]{0.45\columnwidth} 
        \includegraphics[width=\textwidth]{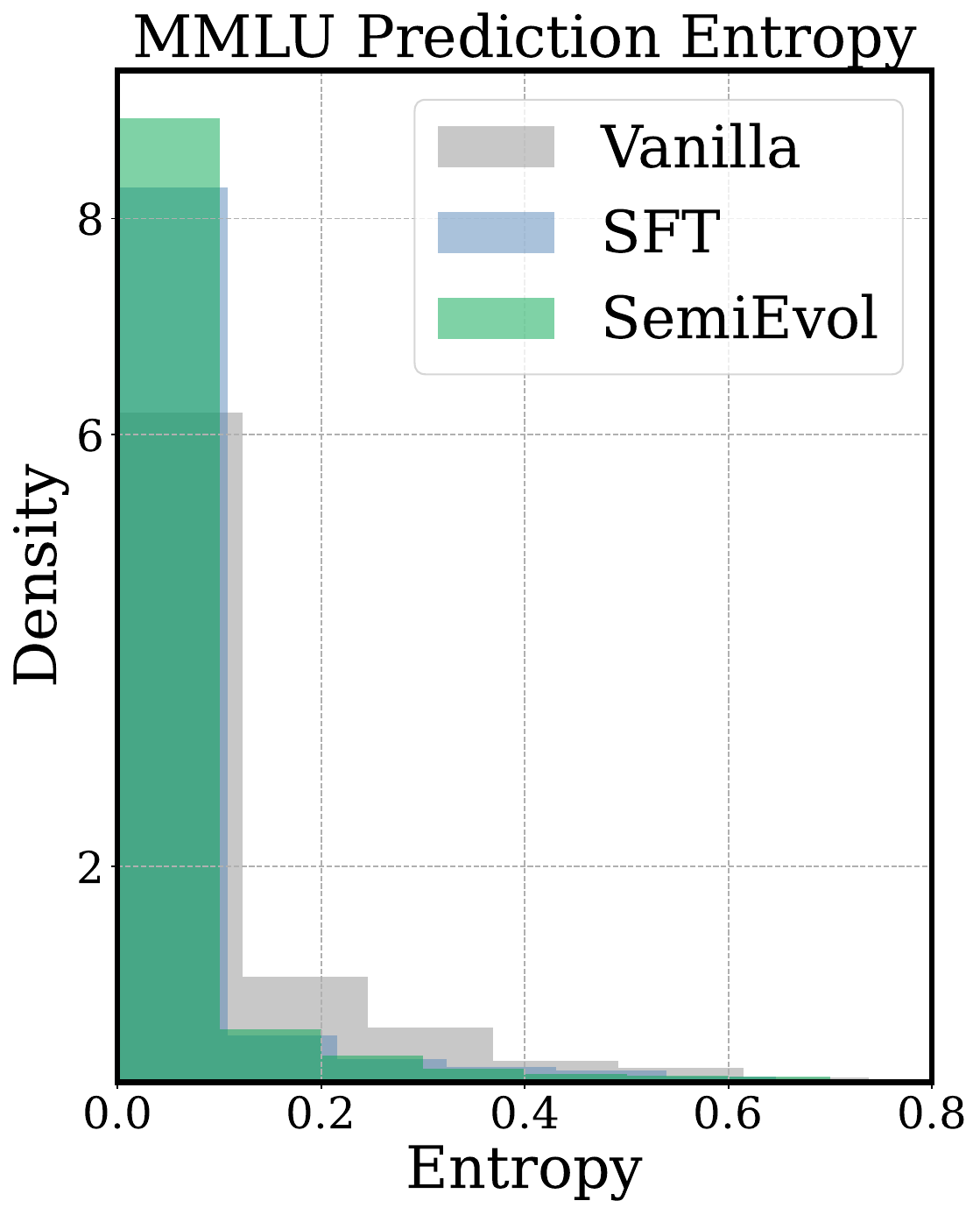}
    \end{subfigure}
    \hfill
    \begin{subfigure}[b]{0.45\columnwidth}
        \includegraphics[width=\textwidth]{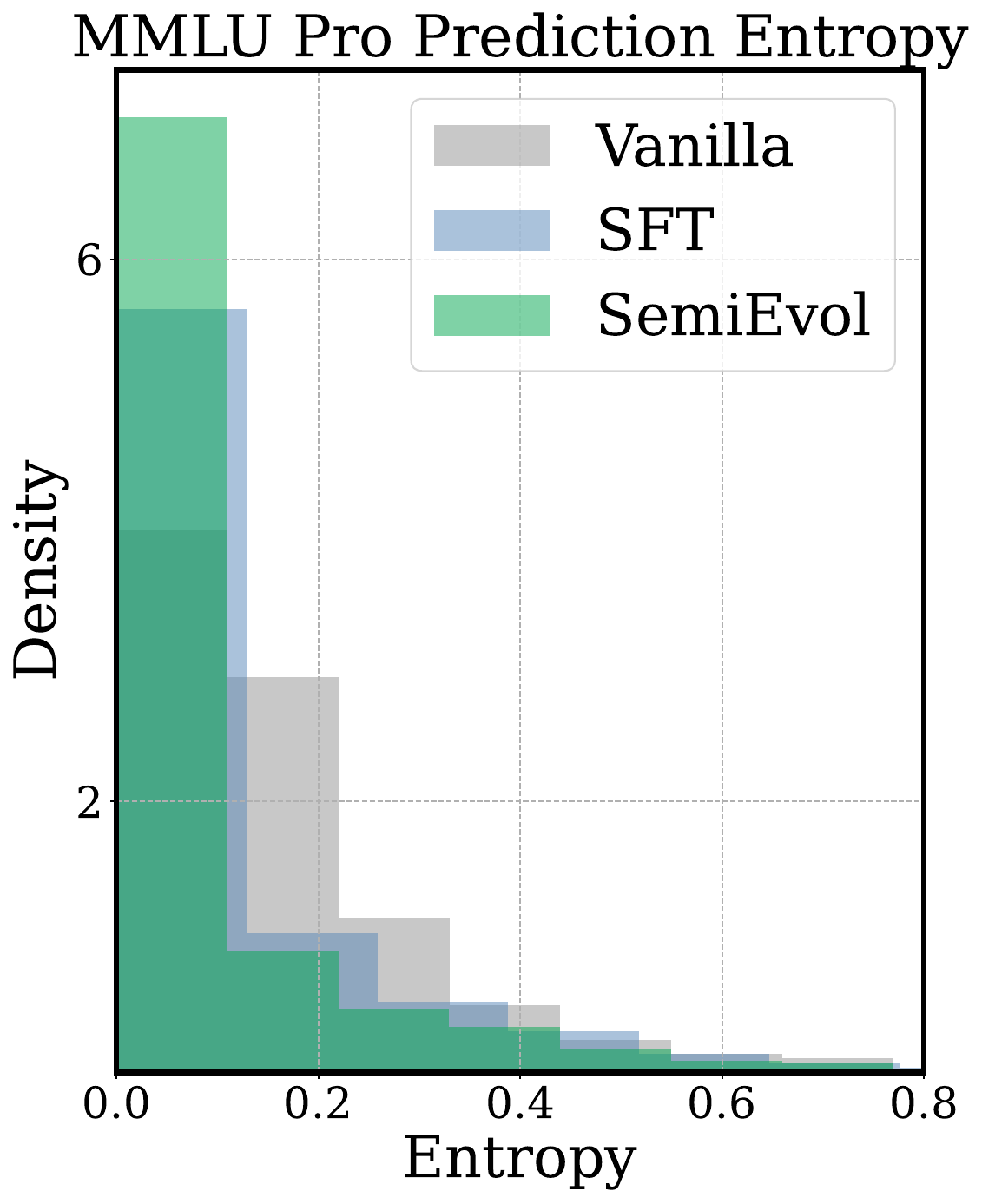} 
    \end{subfigure}
    \caption{\textbf{Entropy distribution} indicates \method{} can enhanced response confidence. Lower entropy values indicate more confident predictions.}
    \label{fig:entropy_analysis} 
\end{figure}

\subsubsection{Sensitivity Analysis}
\label{sec:sen_anal}
We analyze the number of collaborating models~($n$) and the data selection ratio~($\theta$), with results illustrated in Figure~\ref{fig:sensitive}. From the results, we have the following observations. \textbf{(1)} Our method demonstrates robust performance across various settings, indicating low sensitivity to these parameters. \textbf{(2)} Model accuracy generally increases with $n$, as more collaborating LLMs enhance prediction accuracy. However, this also introduces additional computational overhead. We chose $n=4$ as the default. \textbf{(3)} Accuracy initially increases with $\theta$ but subsequently decreases, suggesting that introducing excessively noisy data is detrimental to model evolution. Consequently, we empirically set $\theta=50\%$ as the default value. It is noteworthy that we did not conduct extensive hyperparameter searches, as our primary focus was on validating the overall framework's effectiveness.

\subsubsection{Response Entropy Analysis}
We present the entropy distribution of different methods on the test set, as illustrated in Figure~\ref{fig:entropy_analysis}. Lower entropy indicates more confident responses. Compared to the Vanilla and SFT model, \method{} demonstrates a significant improvement in response confidence. This observation substantiates the effectiveness of \method{} in producing more decisive and assured outputs.
This signifies that \method{} not only improves accuracy but also enhances the model's ability to generate more confident and reliable responses.

\begin{table*}[t]
\centering
\renewcommand{\arraystretch}{1.2}
\setlength{\tabcolsep}{12pt}
\resizebox{.85\linewidth}{!}{
\begin{tabular}{lcccccccc}
\toprule
\textbf{Base Model} & \multicolumn{4}{c}{\textbf{MMLU}~($\gD_{unlabeled}$ \,/\, $\gD_{labled}$)} & \multicolumn{4}{c}{\textbf{MMLU-Pro}~($\gD_{unlabeled}$ \,/\, $\gD_{labled}$)} \\
\cmidrule(lr){2-5} \cmidrule(lr){6-9}
 & \textbf{50\%} & \textbf{100\%} & \textbf{200\%} & \textbf{300\%} & \textbf{50\%} & \textbf{100\%} & \textbf{200\%} & \textbf{300\%} \\
\midrule
GPT-4o mini & 78.2 & 78.6 & 79.3 & 79.9 & 58.9 & 59.5 & 60.1 & 60.8 \\
Llama3.1-8B & 67.6 & 67.9 & 68.6 & 68.8 & 49.8 & 49.8 & 50.0 & 50.3 \\
\bottomrule
\end{tabular}
}
\caption{\textbf{Performance of continuous evolution} with varying amounts of unlabeled data.
}
\label{tab:continuous_evolution}
\end{table*}

\begin{figure}[t] 
    \centering 
    \begin{subfigure}[b]{0.45\columnwidth} 
        \includegraphics[width=\textwidth]{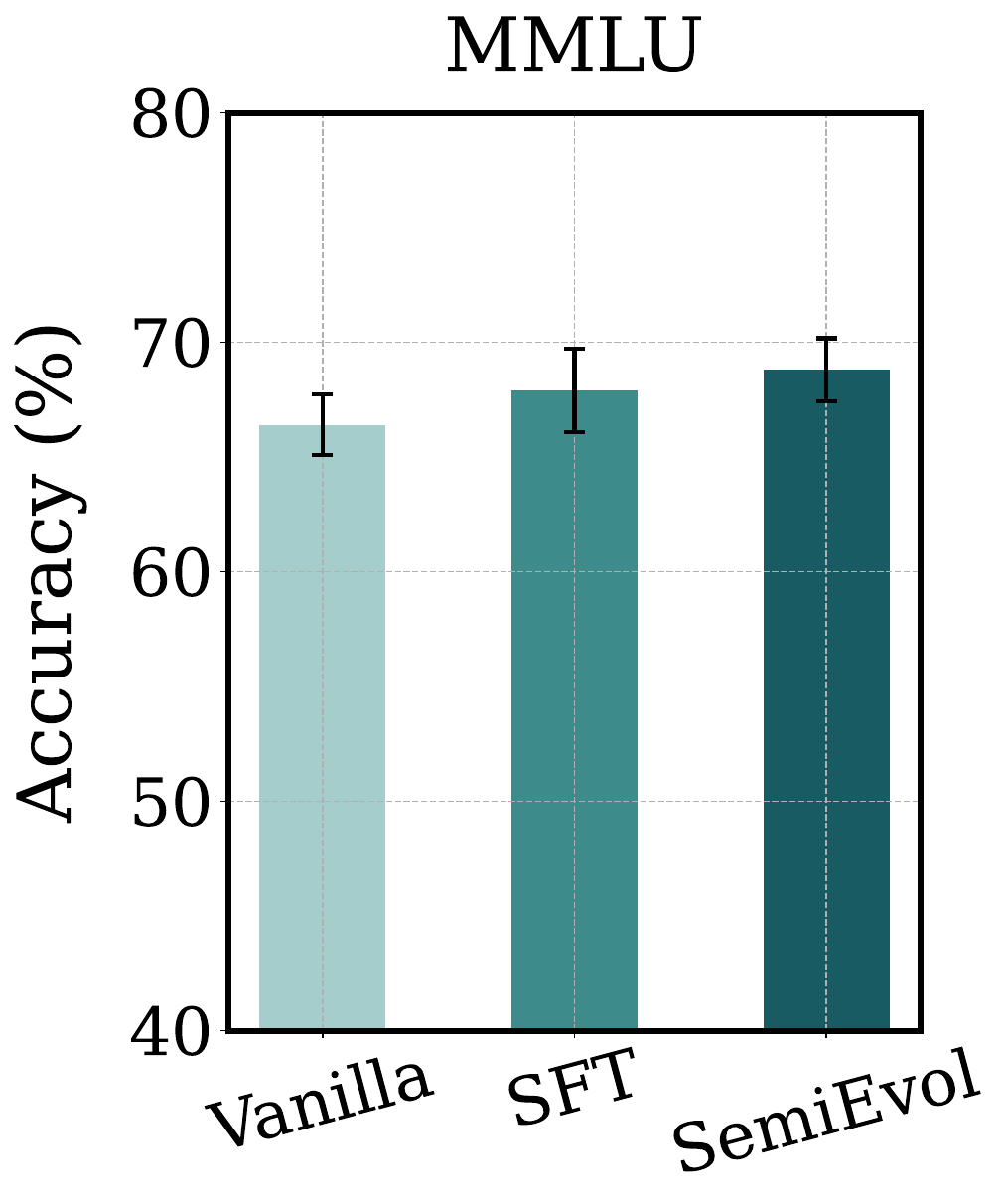}
    \end{subfigure}
    \hfill
    \begin{subfigure}[b]{0.45\columnwidth}
        \includegraphics[width=\textwidth]{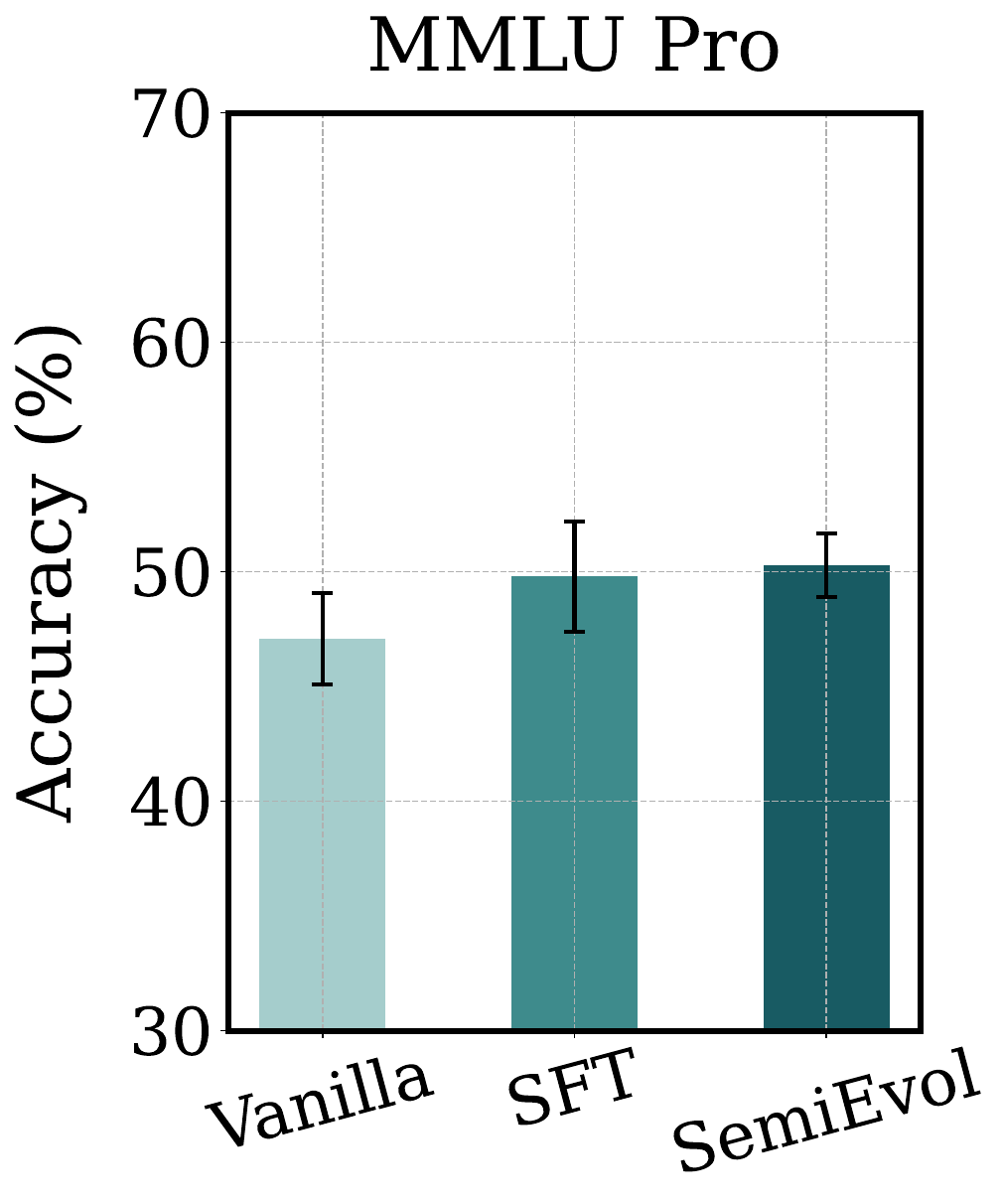} 
    \end{subfigure}
    \caption{\textbf{Stability analysis} via mean performance and standard deviation across multiple inference prompts.}
    \label{fig:stable} 
\end{figure}

\begin{figure}[t] 
    \centering 
    \includegraphics[width=0.8\linewidth]{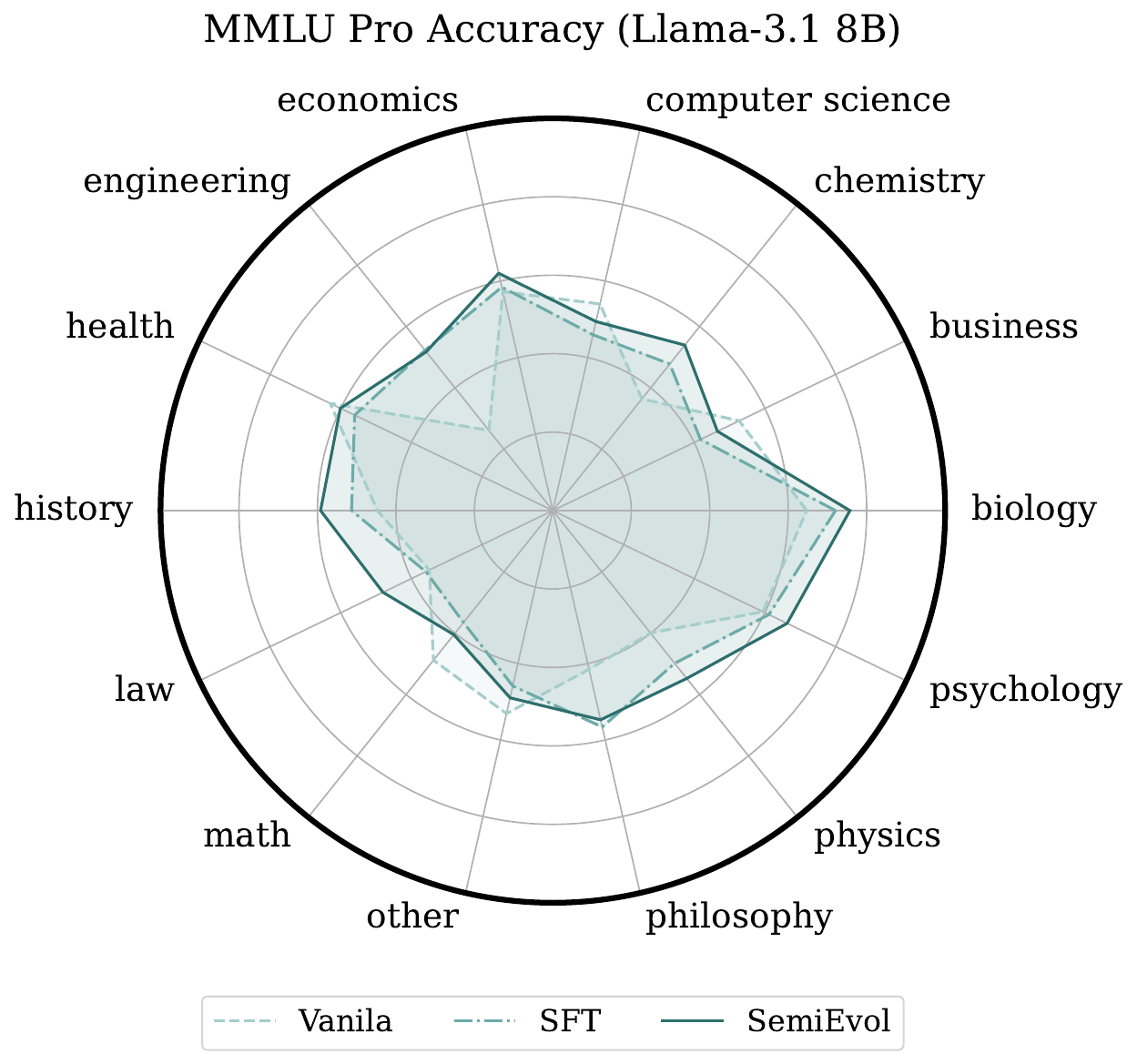}
    \caption{\textbf{Category-wise performance} of \method{}.}
    \label{fig:mmlu_pro_radar} 
\end{figure}

\subsubsection{Category-wise Performance Analysis}
We conducted an in-depth investigation into the differential impact of \method{} across various categories in MMLU-Pro, as illustrated in Figure~\ref{fig:mmlu_pro_radar}. We find that \textbf{(1)} \method{} demonstrates enhanced performance across the majority of domains compared to both SFT and Vanilla approaches. This broad-spectrum improvement underscores the method's versatility and effectiveness across diverse subject areas.
\textbf{(2)} \method{} achieves substantial gains in specific fields such as Law, Engineering, and Philosophy. This notable improvement suggests that knowledge in these domains is underrepresented in common knowledge bases, highlighting the necessity for targeted adaptation.

\subsubsection{Stability Analysis}

We evaluate the inference stability of different models by utilizing diverse prompts. Specifically, we employed GPT-4o to rephrase the instructions and conducted $5$ tests on each model, reporting the average performance and standard deviation. As illustrated in Figure~\ref{fig:stable}, changing the inference prompts had minimal impact on the various models. Notably, \method{} even demonstrated a slight improvement in model stability.

\subsubsection{Discussion on Continuous Evolution}\label{sec:data-ratio}

In real-world scenarios, unlabeled data often accumulates continuously, altering the ratio between labeled and unlabeled data.
Table~\ref{tab:continuous_evolution} illustrates the impact of various data proportions on \method{}'s performance. As illustrated, model performance consistently improves with an increase in unsupervised data across different base models. This validates \method{}'s effectiveness in addressing real-world scenarios, where model performance in specific domains can be progressively enhanced as more unsupervised data accumulates. 

\begin{figure}[t] 
    \centering 
    \includegraphics[width=0.75\linewidth]{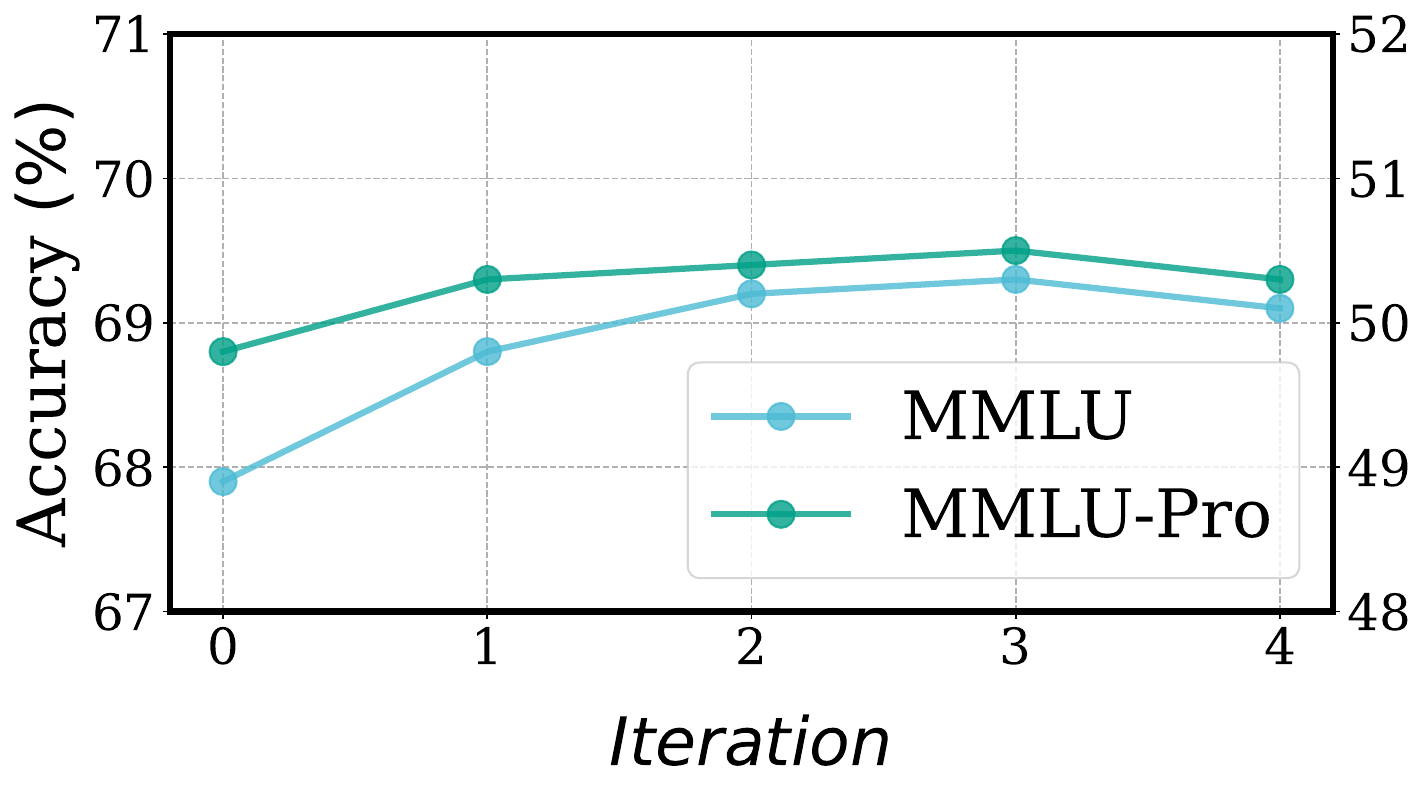}
    \caption{\textbf{Iterative evolution performance}, each iteration means perform a round of \method{}.}
    \label{fig:iter-evol} 
\end{figure}

\subsubsection{Discussion on Iterative Evolution}
We verify the model's iterative evolution capability, as illustrated in Figure~\ref{fig:iter-evol}. After applying \method{}, we utilized the labeled data and pseudo-response data as new labeled data, initiating a fresh round of \method{} on the previously filtered unlabeled data. By the fourth iteration, we had utilized most of the unlabeled data, resulting in further performance improvements in the target scenario. 
This iterative evolution capability further demonstrates the practicality of \method{}.


\section{Related Work}

\subsection{Data Engineering for SFT}

With the rapid advancement of Large Language Models~(LLMs)~\cite{zhao2023survey}, researchers have discovered that employing suitable data for Supervised Fine-Tuning~(SFT) can enhance model performance on downstream tasks~\cite{taori2023stanford,longpre2023flan,hou2024protransformer,luo2024robustft,jiang2024ragraph}. 
Some researchers focus on data selection~\cite{bhatt2024experimental,parkar2024selectllm,xia2024less,bukharin2023data}, aiming to improve data quality to boost model effectiveness within limited training budgets. Others concentrate on data synthesis~\cite{mukherjee2023orca,chung2024scaling,honovich2022unnatural,cheng2023adapting}, attempting to enhance models' instruction-following capabilities through synthesized instruction data. 
Researchers also shifted their focus to model self-evolution~\cite{tao2024survey,madsen2024self}. These include self-instruction~\cite{wang2022self} and self-play~\cite{chen2024self} techniques that enable models to acquire task-specific capabilities without extensive annotated data. 
Complementary to these approaches, \method{} focuses on LLMs' ability to continuously evolve in real-world \textbf{semi-supervised fine-tuning~(SemiFT)} scenarios, relying solely on their inherent capabilities. It effectively utilizes small amounts of labeled data to improve model evolution performance.

\subsection{Semi-supervised Learning}
Semi-supervised learning aims to reduce the annotation cost during model training~\cite{zhu2005semi,tarvainen2017mean,ju2024survey,yang2024poisoning,feng2024bioactivity}, which has received increasing attention in various fields such as text classification~\cite{duarte2023review,thangaraj2018text,linmei2019heterogeneous} and neural machine translation~\cite{cheng-etal-2016-semi,pham2023semi}. Current semi-supervised learning approaches can be mainly divided into two types, \ie, pseudo-labeling~\cite{lee2013pseudo} and consistency regularization~\cite{sohn2020fixmatch,berthelot2019mixmatch}. Pseudo-labeling approaches usually add extra unlabeled data into the labeled dataset by leveraging the labels predicted by the model. Recent studies attempt different techniques to enhance pseudo-labeling such as considering adaptive thresholds~\cite{zhang2024self,rhee2019efficient} and class imbalance~\cite{luo2024gala,wang2024delta}. In contrast, consistency regularization aims to encourage the consistency of predictions under different perturbations. However, these approaches focus on classification problems~\cite{shi2023rethinking}, which cannot be applied to LLM fine-tuning. To tackle this issue, we propose a new framework \method{} in a propagate-and-select manner for LLM adaptation.

\section{Conclusion}

We for the first time investigate the practical challenge of utilizing hybird-data~(\ie, both labeled and unlabeled data) to enhance LLMs performance in specific scenarios. We designed a bi-level framework \method{} for knowledge \textit{propagation-and-selection}. This framework leverages in-weight and in-context knowledge propagation from labeled data, while employing collaborative learning and adaptive selection to generate high-quality pseudo-responses. We validated \method{}'s efficacy on both general and domain-specific datasets, conducting a detailed analysis of the improvements it yields. Furthermore, we demonstrated \method{}'s capability for continuous iterative evolution, which plays a crucial role in enhancing LLMs' effectiveness in real-world applications.

\section*{Limitations}

One limitation of our work is that due to the limit of computational resources, we do not evaluate our framework on more LLMs such as GPT-4o and Llama3.1 70B. In future work, we will attempt to incorporate our framework into these LLMs. Moreover, although our framework is evaluated on various benchmark datasets, we do not involve more complicated domains which require more scientific knowledge. To solve this, we will extend our framework to more advanced scientific domains such as genomics analysis.



\section*{Acknowledgement}

This paper is partially supported by the National Key Research and Development Program of China with Grant No.
2023YFC3341203 as well as the National Natural Science
Foundation of China with Grant Numbers 62276002 and 62306014.


\bibliography{custom}


\newpage

\appendix
\label{appendix}

\section{Algorithm}\label{sup:algo}

In Algorithm~\ref{alg1}, we present the comprehensive algorithmic process of \method{}, which incorporates a bi-level framework for knowledge propagation and selection. This process ultimately yields the evolved model, $\gM_{evol}$.

\begin{algorithm}[h]
    \caption{Algorithm of \method{}}\label{alg1}
    \label{alg:algorithm}
\begin{flushleft}
\textbf{Require}: Labeled data $\gD_{labeled}$, Unlabeled data $\gD_{unlabeled}$, LLM $\gM$;\\
\textbf{Ensure}: Evolved LLM $\gM_{evol}$; 
\end{flushleft}
    \begin{algorithmic}[1] 
    \STATE \textit{// In-Weight Knowledge Propagation} \\
    \STATE Fine-tune $\gM$ on $\gD_{labeled}$, obtain $\gM_{warm}$;\\
    \STATE \textit{// Collaborative Learning}\\
    \FOR{m = 1, $\cdots$, $n$}
    \STATE \textit{// In-Context Propagation}\\
    \STATE Get the prediction $\left\{y^m_j \right\}$ as Eq.~\ref{eq:multi-pred};
    \ENDFOR
    \STATE Self-Justify to generate pseudo-responses;
    \STATE \textit{// Adaptive Selection} \\
    \STATE Select pseudo-responses with entropy as Eq.~\ref{eq:data-selected}, obtain $\gD_{selected}$;
    \STATE Fine-tune on $\gD_{selected}$, obtain $\gM_{evol}$;\\
    \end{algorithmic}
\end{algorithm}

\section{Experimental Settings}

We will detail the experimental process, including parameter settings, prompt configurations, and computational resource consumption.

\subsection{Parameter Settings}

\paratitle{Inference Process} For commercial methods~(\eg, GPT4o-mini), we utilize API calls. For open-source models~(\eg, Llama3.1 8B), we employ the vLLM~\cite{kwon2023efficient} framework locally for inference. The inference parameters are as Table~\ref{tab:sup:param-infer}.

\begin{table}[h]
\centering
\resizebox{\columnwidth}{!}{
\begin{tabular}{ll}
\toprule
Parameter & Value \\
\midrule
Model Name & Llama3.1 8B Instruct \\
Computing Arch. & NVIDIA A100-80GB \\
Max Sequence Length & 4096 \\
\bottomrule
\end{tabular}
}
\caption{Parameters configuration during inference.}
\label{tab:sup:param-infer}
\end{table}

In the collaborative learning process, we employ $n$ LLMs with diverse configurations for mutual learning, setting their temperature to $1$. These configurations are sampled from the following options: (1) Utilization of $\gM_warm$: $\{0, 1\}$; 
(2) Number of labeled samples referenced for in-context knowledge propagation: $\{0, 1, 2, 3\}$. 

\paratitle{Fine-tuning Process}. We use platform APIs for commercial methods fine-tuning and LlamaFactory~\cite{zheng2024llamafactory} for open-source models fine-tuning. Both of them are fine-tuned for $2$ epochs. Commercial models use default settings~(adaptively configured by OpenAI based on the task). For open-source models, our hyperparameter settings are as Table~\ref{tab:sup:param-finetune}, all of which are the default parameters from LlamaFactory.
\begin{table}[h]
\centering
\resizebox{\columnwidth}{!}{
\begin{tabular}{ll}
\toprule
Parameter & Value \\
\midrule
Model Name & Llama3.1 8B Instruct \\
Computing Arch. & NVIDIA A100-80GB \\
Fine-tuning Type & LoRA \\
Gradient Accumulation Steps & 8 \\
Learning Rate & 1e-4 \\
$\theta$ in Adaptive Selection & $50\%$ \\
\bottomrule
\end{tabular}
}
\caption{Hyperparameter settings for fine-tuning open-source models.}
\label{tab:sup:param-finetune}
\end{table}

\subsection{Instruction Settings}

We employ concise instructions for inference as shown in Table~\ref{tab:sup:instruction}. During this process, we present questions to the LLM and elicit responses.

\begin{table}[h]
\centering
\begin{tabularx}{\columnwidth}{X}
\toprule
\textbf{Instruction Template} \\
\midrule
\underline{Multiple Choice} \\
Answer the multiple-choice question.\\
Your response should be in the format: 'Answer: LETTER' (without quotes).\\
\\
Question: \\
\{question\}\\
\\
Options:\\
\{options\} \\
\\
\midrule
\underline{Free-from QA} \\
Answer the following question. \\
Output the value in the format: 
'Answer: VALUE' (without quotes).\\
\\
Question: \\
\{question\}\\
\\
Options:\\
\{options\} \\
\\
Provide your answer on a new line after 'Answer:', without using a \text{\textbackslash boxed} command. \\ 
\midrule
\bottomrule
\end{tabularx}
\caption{Instruction templates for different types of questions.}
\label{tab:sup:instruction}
\end{table}

We extract answers using regular expression matching. For character-based answers, we check for exact matches. For numerical answers, we assess whether they fall within an acceptable error margin (maximum error of 1e-2).

During the \method{} process, we require additional instructions for tasks such as self-justification and in-context knowledge propagation. For these supplementary commands, we provide instruction templates in a table for reference.

\begin{table}[h]
\centering
\begin{tabularx}{\columnwidth}{X}
\toprule
\textbf{Instruction Template} \\
\midrule
\underline{Self-Justify} \\
Here are the multiple answers to the question.\\
Please consider them thoroughly and give me the 
correct answer. Your response should be in the following format: \\
'Answer: LETTER' (without quotes).\\
\\
Question: \\
\{question\} \\ 
 \\
Options:
\{options\} \\
\\
Multiple Answers: \\
\{answers\} \\
\\ 
Now, please give me the final correct answer:\\
\midrule
\underline{In-context Knowledge Propagation} \\
You are an expert in the question answering. Below are some examples of questions and their corresponding answer.\\
\\
\{reference\} \\
\\
\midrule
\bottomrule
\end{tabularx}
\caption{Instruction templates for \method{} process.}
\label{tab:sup:instruction-semievol}
\end{table}

\end{document}